\documentclass[journal,twoside,web]{ieeecolor}
\usepackage{generic}
\usepackage{cite}
\usepackage{amsmath,amssymb,amsfonts}
\usepackage{algorithmic}
\usepackage{graphicx}
\usepackage{algorithm,algorithmic}
\usepackage[
    colorlinks=true,
    linkcolor=blue,
    citecolor=blue,
    urlcolor=blue
]{hyperref}
\hypersetup{hidelinks=true}
\usepackage{textcomp}
\usepackage{booktabs}
\usepackage{multirow}
\usepackage{adjustbox}
\usepackage{booktabs}
\usepackage{makecell}
\usepackage{subcaption}
\usepackage{url}
\usepackage[table]{xcolor}
\providecommand{\refname}{References}
\def\BibTeX{{\rm B\kern-.05em{\sc i\kern-.025em b}\kern-.08em
    T\kern-.1667em\lower.7ex\hbox{E}\kern-.125emX}}
\begin{document}
\title{From Image Interpretation to Clinical Reasoning: Upstream Physician-Context-Aware Multimodal Learning with Causal Reinforcement Learning}

\author{
Jialu~Pi,
Yanan~Ma,
Weijie~Chen,
Owen~Crystal,
Shubham~Trivedi,
Stephen~Xie,
Anna~Silverman,
Matthew~Stib,
Chadi~Ayoub,
Reza~Arsanjani,
and~Imon~Banerjee
\thanks{
Yanan Ma, Weijie Chen, Owen Crystal, Shubham Trivedi, Matthew Stib are with Dept. of Radiology, Mayo Clinic, Phoenix, USA.
}
\thanks{
Stephen Xie, Anna Silverman are with Mayo Clinic, Phoenix, USA.
}
\thanks{
Chadi Ayoub, and Reza Arsanjani are with the Dept. of Cardiology, Mayo Clinic, Phoenix, USA.
}
\thanks{
Jialu Pi, Imon Banerjee is with the Dept. of Radiology, Mayo Clinic, Phoenix, AZ, USA, and with the School of Computing and Augmented Intelligence, Arizona State University, Tempe, USA (e-mail: Pi.Jialu@mayo.edu, jialupi@asu.edu).
}
}

\maketitle
\thispagestyle{empty} 

\begin{abstract}
Major adverse cardiovascular events (MACE) remain the leading cause of mortality worldwide. Opportunistic screening using routinely acquired clinical data offers a scalable approach for identifying high-risk individuals before acute events occur. Although chest X-rays (CXRs) capture latent cardiovascular biomarkers and clinical histories provide complementary patient context, existing medical vision-language models are primarily optimized for radiology interpretation rather than prognostic reasoning. We propose a causal reinforcement learning framework for multimodal clinical reasoning that integrates CXRs and physician-authored clinical histories for opportunistic MACE prediction. The framework introduces (1) a role-decoupled dual-LLM architecture that separates reasoning from risk prediction, (2) a dual-action causal reinforcement learning policy for evidence selection and reasoning optimization, and (3) causal token pruning to learn compact multimodal representations. Evaluated on an internal cohort, an emergency department cohort, and the external MIMIC dataset, the proposed framework consistently outperformed unimodal baselines and state-of-the-art medical vision-language models, achieving AUROCs of 0.720, 0.760, and 0.845, respectively. It also substantially improved reasoning quality, achieving higher GREEN scores and higher expert preference while maintaining robust predictive performance across diverse patient populations.
\end{abstract}


\section{Introduction}
\label{sec:introduction}
\IEEEPARstart{M}{ajor} adverse cardiovascular events (MACE), including myocardial infarction, heart failure, stroke, and cardiovascular death, remain the leading cause of mortality worldwide~\cite{chong2025global}. Early identification of individuals at elevated cardiovascular risk is essential for preventive intervention and reducing long-term morbidity. Opportunistic screening leverages routinely acquired clinical data during standard healthcare encounters to identify high-risk individuals without additional imaging or laboratory testing. Conventional cardiovascular risk models primarily rely on structured clinical variables~\cite{byrne20242023, mathew2025performance, bonassi2008chromosomal} or dedicated cardiac imaging biomarkers~\cite{naghavi2024artificial, zhang2023deepcac}, limiting their scalability and utilization of the rich multimodal information already available in routine care.

Chest radiographs (CXRs) are among the most commonly performed imaging examinations and capture structural cardiopulmonary abnormalities associated with future cardiovascular events. Physician-authored clinical notes accompanying imaging orders provide complementary upstream clinical context, including presenting symptoms, medical history, cardiovascular risk factors, medications, and diagnostic intent. Together, these modalities offer a representation of patient status that more closely reflects the information clinicians synthesize during diagnostic reasoning.

Recent medical vision-language foundation models (e.g. MedCLIP~\cite{wang2022medclip}, BioMedCLIP~\cite{zhang2023biomedclip}, LLaVA-Med~\cite{cli2023llava}, and CheXagent~\cite{chen2024chexagent}) have demonstrated remarkable success in jointly modeling medical images and radiology reports for image interpretation, report generation, and visual question answering. However, these models are predominantly optimized using image--report alignment and next-token prediction objectives, relying on downstream radiology interpretations or structured clinical variables rather than the raw clinical narratives that motivate diagnostic evaluation. Consequently, they often fail to capture the causal relationships between patient presentation, imaging findings, and subsequent cardiovascular outcomes, limiting their ability to perform clinically grounded reasoning for predictive tasks~\cite{guo2025deepseek}.

Reinforcement learning (RL) provides a promising mechanism for aligning multimodal reasoning with downstream clinical objectives~\cite{jiang2026co}. Rather than optimizing only language generation, causal RL treats reasoning as a sequential decision-making process, allowing multimodal large language models (MLLM) to generate reasoning trajectories conditioned on imaging and physician-documented clinical context while receiving feedback based on prediction accuracy and causal consistency. Such optimization encourages the model to identify clinically meaningful evidence, suppress spurious associations, and produce reasoning that is both interpretable and predictive of future patient outcomes.

In this work, we propose a causal reinforcement learning framework for multimodal clinical reasoning with LLM that integrates chest X-rays and physician-authored clinical histories for opportunistic MACE prediction. Our framework introduces three key technical innovations: (1) a decoupled multimodal LM architecture that separates reasoning generation from outcome prediction using dual language models, (2) a dual-action causal reinforcement learning policy that jointly learns clinically relevant token selection and reasoning optimization through causal interventions, and (3) a causal token-pruning strategy that focuses the model on important tokens and produces compact causal embeddings through a multi-objective RL reward balancing predictive performance and token density. We evaluate the proposed framework across three heterogeneous cohorts, including an internal hospital cohort, an emergency department cohort, and the publicly available MIMIC dataset. 

\section{Background}
Recent medical foundation models have predominantly focused on learning joint representations between medical images and corresponding radiology reports through large-scale vision–language pretraining~\cite{lee2024llm, huang2024chest}. Open-source multimodal medical models, including MedCLIP~\cite{wang2022medclip}, BioMedCLIP~\cite{zhang2023biomedclip}, LLaVA-Med~\cite{cli2023llava}, and CheXagent~\cite{chen2024chexagent}, have further advanced medical image understanding by integrating visual encoders with large language models for tasks like chest X-ray interpretation, visual question answering, and automated report generation. These approaches typically optimize image–text alignment objectives using radiology reports or curated medical captions, enabling strong performance in image interpretation, report generation, and cross-modal retrieval~\cite{lee2025cxr}. However, such representations primarily capture the association between imaging findings and their downstream expert interpretation rather than the upstream clinical reasoning process that drives diagnostic decision-making. Radiology reports provide a focused description of observed imaging abnormalities but often lack critical contextual information, including presenting symptoms, medical history, physical examination findings, and physician diagnostic hypotheses. Although recent studies have explored prompt-guided generation of structured radiology reports by coupling medical vision encoders with pretrained large language models~\cite{li2024prompt}, existing multimodal medical models remain susceptible to hallucination, spurious correlations, and inconsistent reasoning trajectories~\cite{das2025trustworthy, chu2025reducing}. These limitations highlight the need for multimodal frameworks that incorporate upstream clinical context and develop more reliable, transparent, and clinically grounded reasoning mechanisms for high-stakes medical applications.

Incorporating upstream clinical context is essential for developing AI systems that move beyond image recognition and report generation toward comprehensive clinical reasoning. However, existing multimodal medical AI approaches face several fundamental challenges in achieving this goal~\cite{heiman2025factchexcker, zhang2024uncovering}. First, most vision–language foundation models rely on highly aligned image–report pairs, where both modalities describe overlapping diagnostic information. Although this alignment facilitates representation learning and text generation, it does not necessarily optimize the model for clinical reasoning tasks, which require establishing meaningful associations between multimodal evidence and targeted clinical outcomes or decisions. Second, clinical reasoning requires integration of heterogeneous and incomplete information, including presenting symptoms, medical history, physical examination findings, temporal disease trajectories, and physician diagnostic hypotheses. Clinical notes introduce additional complexity due to variable documentation styles, missing information, and implicit reasoning patterns. Therefore, developing multimodal AI systems that align imaging and physician-authored narratives with clinically meaningful prediction objectives, while enabling reliable and interpretable reasoning, remains a critical unresolved challenge.

Recent advances in multimodal LLM architectures, such as VILA~\cite{lin2024vila}, NVILA~\cite{liu2025nvila}, have introduced scalable approaches for transforming heterogeneous inputs into unified multimodal token sequences. Unlike conventional image-text alignment approaches optimized primarily for description generation, these architectures enable cross-modal attention and reasoning over visual and linguistic compressed token representations. In healthcare, this paradigm provides a foundation for integrating imaging data with upstream clinical narratives, where physician notes encode symptoms, medical history, and diagnostic intent. However, effective clinical reasoning requires additional alignment between multimodal token representations and downstream clinical objectives rather than solely optimizing language generation. Current multimodal LLMs are primarily trained to maximize the likelihood of generating plausible textual outputs given multimodal inputs. While this enables strong language generation capabilities, it does not explicitly optimize whether the generated reasoning process leads to accurate clinical predictions, appropriate diagnostic hypotheses, or clinically meaningful decisions.

Reinforcement learning (RL) has emerged as a promising approach for aligning foundation models (FMs), including large language models (LLMs) and vision-language models (VLMs), with task-specific objectives~\cite{li2023remax}. However, reward engineering remains a major challenge, particularly in complex domains where desired behaviors cannot be easily specified. Recent studies have explored leveraging FMs as synthetic feedback sources, either by generating reward functions or providing multimodal evaluation signals~\cite{ma2024eureka, venuto2024code}. While these approaches enable scalable supervision, directly generated rewards are often noisy and high-variance.

More recent approaches have adopted preference-based reinforcement learning, where FMs act as evaluators to generate trajectory-level preference signals for reward model training~\cite{wang2026primt}. These methods have demonstrated improved stability by using FMs as critics rather than direct reward generators. However, existing FM-guided RL frameworks have primarily focused on task-oriented domains and do not address the challenges of clinical reasoning, where effective decision-making requires integrating heterogeneous multimodal evidence, temporal context, and causal relationships between clinical observations and outcomes.

\section{Methods} 
\subsection{Overall Framework}
\begin{figure}[htb!]
    \centering

    \begin{subfigure}[b]{0.48\textwidth}
        \centering
        \includegraphics[width=\textwidth]{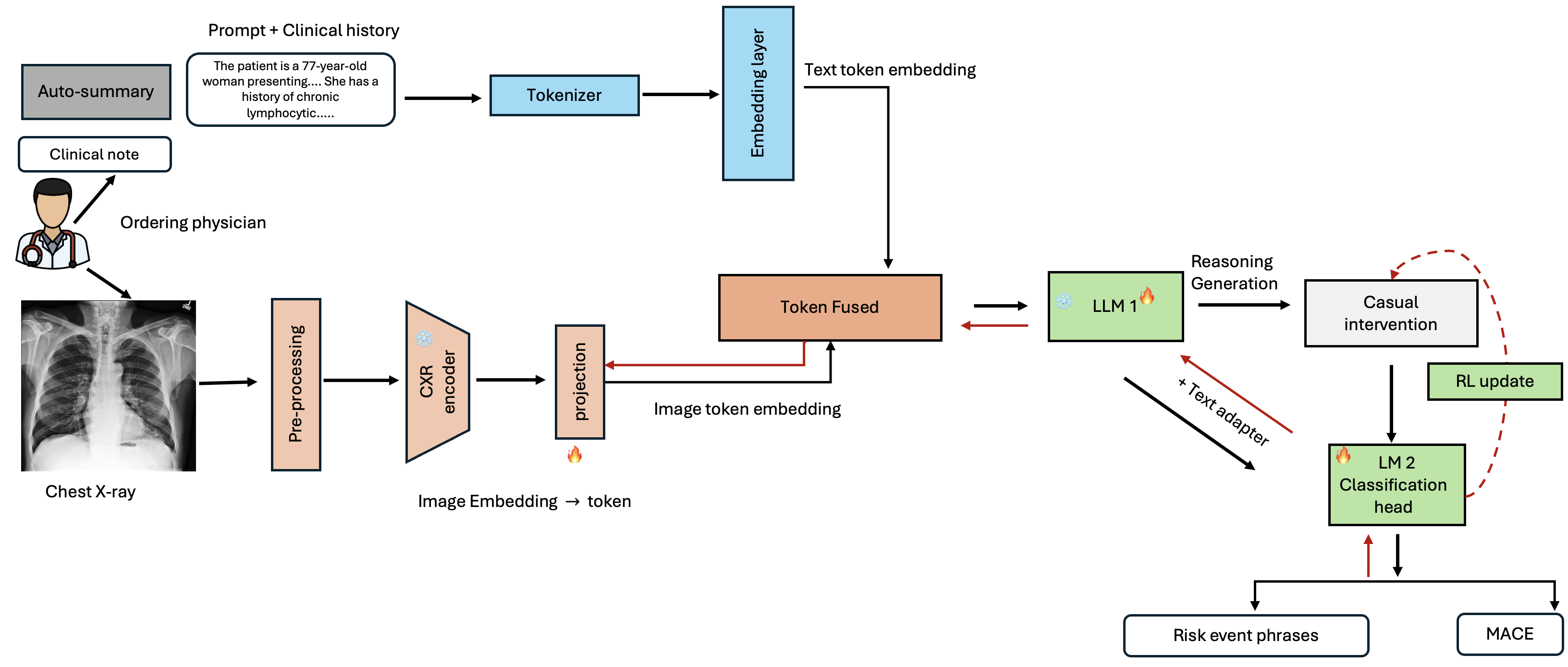}
        \caption{}
        \label{fig:sub1}
    \end{subfigure}
    \hfill
    \begin{subfigure}[b]{0.48\textwidth}
        \centering
        \includegraphics[width=\textwidth]{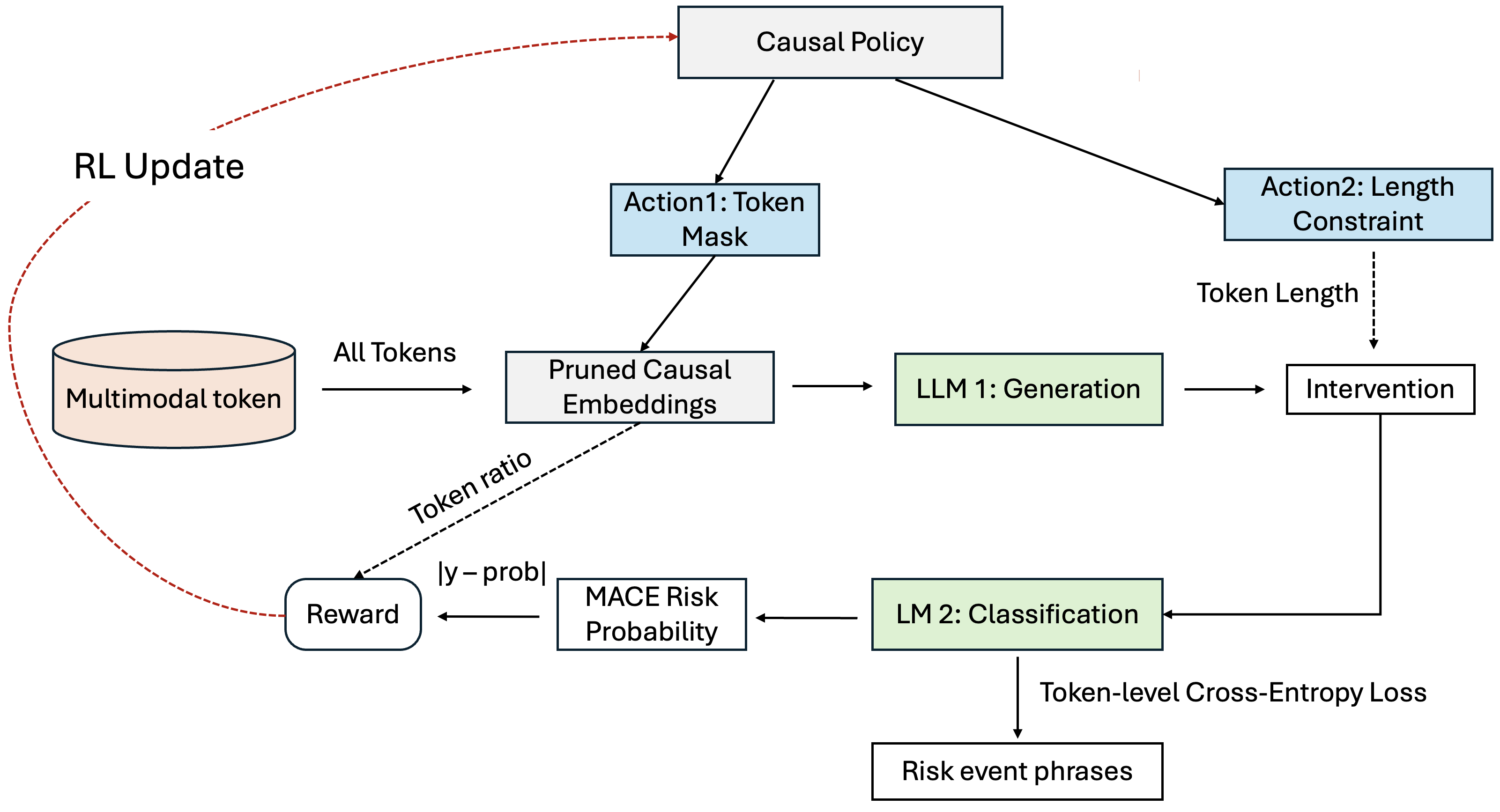}
        \caption{}
        \label{fig:sub2}
    \end{subfigure}
    \caption{\footnotesize Proposed causal reinforcement learning framework with a multimodal LLM. (a) End-to-end training pipeline integrating multimodal inputs, multimodal token generation, causal reasoning, and reinforcement learning. Red arrows indicate gradient backpropagation and parameter updates. (b) Causal policy optimization, where reinforcement learning refines the reasoning policy using causal feedback to improve reasoning consistency and align generated reasoning with targeted clinical prediction objectives. LLM denotes large language model, and LM denotes language model.}
    \label{fig:main}
\end{figure}

We propose a Foundation Model (FM)-guided causal reinforcement learning (RL) framework for multimodal clinical reasoning (Fig.~\ref{fig:main}). The framework optimizes prediction-oriented reasoning by allowing a MLLM to generate reasoning trajectories and receive clinically informed feedback. Unlike conventional approaches that optimize language generation objectives alone, our framework aligns multimodal representations from medical images and clinical text with clinical decision objectives to improve prediction reliability and reasoning fidelity. To mitigate the competing optimization objectives between multimodal reasoning and risk prediction, we employ a role-decoupled dual-LM architecture. The first large LM serves as a multimodal reasoning module that integrates chest X-ray (CXR) images and clinical information to generate clinically relevant risk summaries. The second LM performs downstream risk classification and outcome phrase generation. This separation reduces task interference and enables more effective optimization of both reasoning generation and predictive performance.



For each patient (i), the input consists of a chest X-ray image ($I_i$) and associated clinical history and demographic information summarize from ordering physician note ($C_i$). The prediction target ($y_i \in {0,1}$) denotes the occurrence of a MACE within the predefined prediction horizon. For positive cases, the associated outcomes include non-hemorrhagic stroke, myocardial infarction, acute heart failure, all-cause mortality, and unplanned cardiac-related hospitalization. The objectives are to estimate the MACE probability ($p(y_i|I_i,C_i)$), generate multimodal clinical reasoning conditioned on ($I_i$) and ($C_i$), and produce a concise outcome description consistent with the predicted risk.

\subsection{Role-Decoupled Dual-LM Framework (Stage-I Supervised Training)} 

A single multimodal decoder must simultaneously perform visual-textual integration, clinical reasoning generation, risk classification, and outcome generation, which may introduce competing optimization objectives. We therefore adopt a role-decoupled dual-LM architecture, where LLM1 ($\theta_1$) integrates image and clinical information to generate concise cardiovascular risk reasoning, and LM2 ($\theta_2$) - BioGPT performs risk classification and outcome-phrase generation. This design separates reasoning and prediction while allowing BioGPT to share its backbone for classification and structured generation.

\subsubsection{Multimodal Reasoning Model}
Given a chest X-ray image \(I_i\), clinical history \(C_i\), and task prompt, LLM1 ($\theta_1$) generates a free-text cardiovascular-risk summary:
\(
R_i = G_{\theta_1}(I_i,C_i).
\) Because this summary is generated through discrete token sampling and subsequently retokenized by LM2 ($\theta_2$), gradients from the downstream classifier cannot propagate through \(R_i\).
To provide a differentiable pathway from the original multimodal input, we also extract the contextualized hidden state of the last valid VLM input token. This representation is projected into the $\theta_2$ hidden space using a trainable linear adapter:
\(
\mathbf v_i =
A_{\psi}\!\left[
\operatorname{Pool}\!\left(F_{\theta_1}(I_i,C_i)\right)
\right],
\) where \(F_{\theta_1}\) denotes the VLM forward pass, \(\operatorname{Pool}(\cdot)\) selects the last valid token representation, and \(A_{\psi}\) is the multimodal feature adapter. The downstream model therefore receives complementary information from both the generated clinical reasoning \(R_i\) and the direct multimodal representation \(\mathbf v_i\).

\subsubsection{Risk Classification and Event Generation}

LM2 ($\theta_2$) encodes the VLM-generated reasoning \(R_i\), and its last valid hidden state, \(\mathbf h_i\), serves as the textual representation of the study. This representation is fused with the projected multimodal feature \(\mathbf v_i\) by element-wise addition where both vectors have the $\theta_2$ hidden dimension. A linear classification head then produces the class-probability vector as: \(
\mathbf p_i=
\operatorname{softmax}
\left(
\mathbf W_{\mathrm{cls}}(\mathbf h_i+\mathbf v_i)
+\mathbf b_{\mathrm{cls}}
\right),
\) where \(\mathbf W_{\mathrm{cls}}\) and \(\mathbf b_{\mathrm{cls}}\) are the trainable weight matrix and bias of the classification head. The component of \(\mathbf p_i\) corresponding to the positive class is interpreted as the predicted probability of a MACE. Classification is optimized using class-weighted cross-entropy to account for class imbalance.

$\theta_2$ is additionally trained to append a short outcome phrase after \(R_i\). The decision gate \(d_i\) is determined by the Multimodal Reasoning Model class predicted by the risk classifier:
\(
d_i=\arg\max_{c\in\{0,1\}} p_{i,c},
\) where \(p_{i,c}\) is the predicted probability that sample \(i\) belongs to class \(c\). \(p_{i,0}\) and \(p_{i,1}\) denote the predicted probabilities of the negative/positive MACE classes respectively. When \(d_i=0\), the target is a fixed sentence indicating non-elevated cardiovascular risk. When \(d_i=1\) and the ground-truth label is positive, the target describes the observed combination of cardiovascular events. When \(d_i=1\) but the ground-truth label is negative (i.e., a false-positive prediction), an event combination is sampled from the empirical distribution of the training set and used as the generation target. The gates are recomputed during epoch-level gating procedure using the classifier’s updated predictions.

Let \(P(R_i,d_i)\) denote the generation prefix containing the reasoning and gate information, and let
\(Q_i=(q_{i,1},\ldots,q_{i,K_i})\) be the target outcome phrase. The generation objective is
\(\mathcal{L}_{\mathrm{gen}}
=
-\frac{1}{\sum_i K_i}
\sum_i\sum_{t=1}^{K_i}
\log p_{\theta_2}
\left(
q_{i,t}\mid P_i,q_{i,<t}
\right).\)
where \(K_i\) is the number of target tokens, \(\theta_2\) denotes the BioGPT parameters, \(q_{i,t}\) represents target token at position \(t\) and \(q_{i,<t}\) is preceding target tokens supplied through teacher forcing.

Loss is computed only on target outcome tokens; tokens belonging to the prefix are masked out. The total supervised objective combines cross-entropy loss for classification and the above generation loss:
\(
    \mathcal{L}_{\mathrm{Stage_I}}
    =
    \lambda_{\mathrm{cls}}\mathcal{L}_{\mathrm{cls}}
    +
    \lambda_{\mathrm{gen}}\mathcal{L}_{\mathrm{gen}},
\)
where \(\lambda_{\mathrm{cls}}\) and \(\lambda_{\mathrm{gen}}\) control the relative contributions of the two tasks.


\subsection{Intervention-Guided Causal Reinforcement Learning (Stage-II Causal RL Training)}

Correct classification alone does not ensure that the generated reasoning relies on concise and relevant multimodal evidence. We therefore learn an intervention policy that removes selected input tokens before reasoning generation and controls the maximum reasoning budget (Fig.~\ref{fig:main}.b). The downstream classification and generation performance under each intervention provide the reinforcement signal. Inspired by the causal policy formulation in~\cite{hwang2024finegrained}, we use ``causal'' to describe interventions on the model's internal information pathway, rather than causal inference about clinical exposures or patient outcomes.

\subsubsection{Quantized Causal Policy State}

For each study, we capture the selected VLM's native fused representation immediately before the language decoder. The policy state includes the projected visual and text embeddings. Let \(\mathbf E_i =
(\mathbf e_{i,1},\ldots,\mathbf e_{i,T_i})\in\mathbb R^{T_i\times d_1}\) denote the VLM’s fused pre-decoder representation for study \(i\), where \(\mathbf e_{i,t}\in\mathbb R^{d_1}\) is the embedding of the \(t\)-th multimodal input token. Together with their attention mask, positional indices, segment labels, and clinical-word groups. The segment labels distinguish system-prompt, image, clinical-text, task-prompt, and other tokens.

The context-aware policy combines each token with the global study context:
\(
    \bar{\mathbf e}_i
    =
    \frac{1}{T_i}\sum_{t=1}^{T_i}\mathbf e_{i,t},
    \qquad
    \mathbf x_{i,t}
    =
    [\mathbf e_{i,t};\bar{\mathbf e}_i],
\)
where \([\cdot;\cdot]\) denotes concatenation. Thus, a token is evaluated using both its own content and the complete multimodal context. The mean embedding \(\bar{\mathbf e}_i\) provides a global summary of the multimodal input. By concatenating this global context with each token embedding, the policy evaluates each token relative to the full study rather than in isolation.

Inspired by fine-grained causal dynamics learning, we quantize the token-context policy states into recurring discrete contexts. 
For each token-level state \(\mathbf x_{i,t}\), an encoder first produces a latent representation as
\(
    \mathbf z^e_{i,t}=f_{\mathrm{enc}}(\mathbf x_{i,t}).
\)
This latent vector is assigned to the nearest prototype in an EMA-updated codebook 
\(\mathcal C=\{\mathbf c_k\}_{k=1}^{K}\):
\(
    k_{i,t}^{*}
    =
    \arg\min_{k}
    \|\mathbf z^e_{i,t}-\mathbf c_k\|_2^2,
    \qquad
    \mathbf z^q_{i,t} =
    \mathbf c_{k_{i,t}^{*}} .
\)
Here, \(k_{i,t}^{*}\) denotes the selected discrete context index, and \(\mathbf z^q_{i,t}\) is the corresponding quantized representation. The codebook is updated by exponential moving averages of the assigned encoder outputs, and a commitment loss later encourages \(\mathbf z^e_{i,t}\) to remain close to its selected prototype \(\mathbf z^q_{i,t}\). This step groups continuous token-context states into a finite set of reusable policy contexts. We use the straight-through estimator to pass gradients through the quantized representation. The EMA codebook is updated from the assigned encoder outputs, and a commitment regularization term  encourages encoder outputs to remain close to their selected codes.

During training, the non-differentiable nearest-neighbor assignment is handled with a straight-through estimator, which allows downstream policy gradients to update the encoder through the quantized representation. The EMA codebook is updated from the assigned encoder outputs, and a commitment regularization term encourages encoder outputs to remain close to their selected codes. A decoder then maps the quantized latent state to a policy feature:
\(
    \mathbf h_{i,t}
    =
    f_{\mathrm{dec}}(\mathbf z^q_{i,t})
=
f_{\mathrm{policy}}(\mathbf x_{i,t}),
\)
The policy uses \(\mathbf h_{i,t}\) to predict evidence-selection actions. Thus, the quantized contexts provide a discrete state representation for deciding which multimodal tokens should be retained for VLM reasoning.


\subsubsection{Policy Actions and Causal Intervention}

The policy first converts each multimodal token embedding into a context-aware policy state
\(\mathbf x_{i,t}\). After encoding, quantization, and decoding, these states produce policy features 
\(\mathbf h_{i,t}\). The joint action is then sampled from the policy conditioned on the decoded 
feature sequence \(\mathbf H_i=(\mathbf h_{i,1},\ldots,\mathbf h_{i,T_i})\):
\(
    a_i=(\mathbf m_i,b_i)\sim\pi_\phi(\cdot\mid\mathbf H_i).
\)
Here, \(\mathbf m_i=(m_{i,1},\ldots,m_{i,T_i})\) is a binary token-retention mask, and \(b_i\) selects 
a reasoning-length bin. The token-retention action is sampled from a Bernoulli distribution, 
\(
    m_{i,t}\sim
    \operatorname{Bernoulli}
    \left(
        \rho_{i,t}
    \right),
\)
where \(f_{\mathrm{keep}}\) is an MLP token-retention head and \(\rho_{i,t} = \sigma(f_{\mathrm{keep}}(\mathbf h_{i,t})).\) The length action is sampled from a 
categorical distribution,
\(
    b_i\sim
    \operatorname{Categorical}
    \left(
        \operatorname{softmax}
        (f_{\mathrm{len}}(\bar{\mathbf h}_i))
    \right),
\)
where \(\bar{\mathbf h}_i\) is the mean of valid decoded token features and \(f_{\mathrm{len}}\) is 
an MLP length-selection head. The decoded policy features \(\mathbf h_{i,t}\) are passed to a token-retention head \(f_{\mathrm{keep}}\), implemented as an MLP, to produce the keep probability for each token. The mean pooled feature \(\bar{\mathbf h}_i\) is passed to a length-selection head \(f_{\mathrm{len}}\), also implemented as an MLP, to choose a reasoning-length bin \(b_i\) from a predefined budget set \(\mathcal B\). The Bernoulli distribution over token-retention decisions also provides an entropy term, which is used during RL optimization to encourage exploration of different evidence masks. During training, hard masks are sampled using a straight-through Gumbel-sigmoid estimator. The selected bin determines the maximum generation budget used by the VLM.

Before rollout, the sampled token-retention mask is converted into an effective mask \(\widehat{\mathbf m}_i\) by a constraint operator. This step ensures that system and task-instruction tokens are always retained, minimum fractions of image and clinical tokens are preserved, and clinical subwords are kept or removed as 
complete words. Tokens beyond the maximum sequence length supported by the policy are also retained. The effective mask is applied by retaining only token embeddings who will be kept, while preserving their original order in the multimodal sequence. The corresponding attention-mask entries and multimodal positional indices are gathered at the same positions. The VLM generates the reasoning text from this masked evidence sequence, using the maximum generation budget selected by the length action. Thus, the token-retention action determines which multimodal evidence can influence reasoning, while the length action controls the maximum reasoning budget.

\subsubsection{Rollout Reward}

After \(G\) rollouts are sampled for study \(i\), each intervention produces a reasoning text 
\(R_i^{(g)}\). BioGPT encodes \(R_i^{(g)}\) together with the unchanged study-level multimodal 
feature, and outputs the MACE-positive probability \(p_i^{(g)}\). Thus, the RL 
intervention modifies the VLM reasoning pathway, while the direct multimodal pathway to the 
classifier remains unchanged. 

For each rollout, three costs are computed:
\(
    c_{\mathrm{err},i}^{(g)}
        = |y_i-p_i^{(g)}|,
    c_{\mathrm{tok},i}^{(g)}
        = \frac{\|\widehat{\mathbf m}_i^{(g)}\|_0}{T_i},
    c_{\mathrm{gen},i}^{(g)}
        = \mathcal L_{\mathrm{gen},i}^{(g)} .
\)
The first cost measures MACE prediction error, and the second measures the proportion of retained 
input tokens. The third cost is the target-token generation loss for the ground-truth outcome phrase, 
computed using the rollout reasoning \(R_i^{(g)}\) as context. In the RL reward, this outcome target 
is determined by the ground-truth MACE label and observed event combination, rather than by the 
classifier-derived generation gate used during supervised training.

The scalar reward is
\[
    r_i^{(g)}
    =
    -w_{\mathrm{err}}c_{\mathrm{err},i}^{(g)}
    -w_{\mathrm{tok}}c_{\mathrm{tok},i}^{(g)}
    -w_{\mathrm{gen}}c_{\mathrm{gen},i}^{(g)}.
\]
A high-reward intervention must therefore preserve evidence needed for MACE classification and outcome generation while removing unnecessary tokens.

\subsubsection{Group-Relative Policy Optimization}

For every study, \(G\) actions are sampled from the rollout policy. Standard GRPO compares each rollout with the other interventions generated for the same study:
\(
    A_i^{(g)}
    =
    \frac{
        r_i^{(g)}-\bar r_i
    }{
        \sigma_i^{r}+\epsilon
    },
\)
where \(\bar r_i\) and \(\sigma_i^{r}\) are the mean and standard deviation of the \(G\) rewards for study \(i\). This patient-specific baseline removes the need for a separately trained value model.

Let \(\phi_{\mathrm{old}}\) denote the rollout policy used to sample the interventions. 
For each rollout, we compute the joint-action importance ratio
The joint action is sampled from the policy conditioned on the decoded policy feature sequence:

\(
    \eta_i^{(g)}
    =
    \frac{
        \pi_{\phi}(a_i^{(g)}\mid \mathbf H_i)
    }{
        \pi_{\phi_{\mathrm{old}}}(a_i^{(g)}\mid \mathbf H_i)
    },
\)
where \(a_i^{(g)}=(\widehat{\mathbf m}_i^{(g)},b_i^{(g)})\) includes the constrained token-retention 
mask and the selected length bin. Let
\(
\bar\eta_i^{(g)}
=
\operatorname{clip}
(\eta_i^{(g)},1-\epsilon_c,1+\epsilon_c).
\)
The clipped GRPO loss is
\[
\mathcal L_{\mathrm{GRPO}}
=
-\frac{1}{NG}
\sum_{i,g}
\min
\left[
\eta_i^{(g)}A_i^{(g)},
\bar\eta_i^{(g)}A_i^{(g)}
\right].
\]
We do not include an explicit KL penalty to a reference language model because the RL policy operates over structured intervention actions rather than language-token generation. Policy drift is instead controlled by clipped importance ratios, entropy regularization, sparsity regularization, and hard evidence-retention constraints.

The complete causal-policy objective $\mathcal{L}_{\mathrm{Stage_{II}}} = \mathcal L_{\mathrm{RL}}$ is:
\[  \mathcal L_{\mathrm{GRPO}}
    +
    \beta_{\mathrm{com}}\mathcal L_{\mathrm{com}}
    +
    \beta_{\mathrm{keep}}\mathcal L_{\mathrm{keep}}
    -
    \beta_H\mathcal H_{\mathrm{Bern}} .
\]
Here, \(\mathcal L_{\mathrm{com}}\) is the commitment loss introduced by the quantized policy state, which encourages encoder outputs to remain close to their assigned codebook vectors. The auxiliary keep-probability loss is defined as
\(
    \mathcal L_{\mathrm{keep}}
    =
    \frac{1}{\sum_i T_i}
    \sum_i\sum_{t=1}^{T_i}
    \rho_{i,t},
\)
where \(\rho_{i,t}\) is the probability of retaining token \(t\) for study \(i\). 
The sparsity term encourages compact evidence selection by penalizing excessive token retention, while the rollout reward uses the effective hard mask after sampling and constraints. The Bernoulli entropy term provides exploration regularization to prevent premature policy collapse. The codebook is updated using exponential moving averages without an additional gradient-based loss. Stage-II causal RL alternates supervised updates with GRPO optimization of the causal policy and uses interventional reasoning from pruned inputs for BioGPT adaptation. During inference, token retention and reasoning length are deterministically selected using learned policy probabilities and applied before final reasoning and classification.

\section{Experimental Setup}
\subsection{Datasets}
\subsubsection{Internal Cohort}
Following Institutional Review Board approval, we collected three internal cohorts from four geographically diverse sites, including retrospective inpatient (24,349 studies from 7,386 patients with a 25.37\% one-year MACE rate), prospective inpatient (8,737 studies from 4,370 patients, with a 35.57\% one-year MACE rate), and retrospective cardiology patient cohorts (10,245 studies from 4,147 patients, with a 51.54\% one-year MACE rate). Each cohort comprised paired chest radiographs and temporally matched clinical notes with one-year MACE outcomes. Patients without sufficient follow-up for outcome ascertainment were excluded. Together, these cohorts represent high-risk cardiovascular populations and were used for model development and internal evaluation. The more detailed demographic characteristics are provided in Table~\ref{tab:demographics}.





\subsubsection{Shift cohort (ED)} To evaluate the model's robustness under distrubution `shift', we constructed an Emergency Department (ED) cohort consisting of patients undergoing chest radiography for any clinical indication. Compared with the internal cohort, this ED cohort with significantly low presence of comorbidities, such as CKD (0.3\%), diabetes (0.9\%) and hypertension (1.5\%), as well as low MACE rate (10.82\%) within one year (Table~\ref{tab:demographics}). 

\subsubsection{External-Shift cohort (MIMIC)} To further assess external generalization, we evaluated the model on a publicly available MIMICIV eICU data~\cite{johnson2020mimic}. We randomly selected 5,000 study of 4,616 patients who underwent inpatient CXR imaging for any clinical indication were matched with ordering clinical notes. We curated the one year MACE outcome by parsing the clinic notes, ICD codes and inpatient mortality data and 1 year MACE rate was 21.82\%(Table~\ref{tab:demographics}).

\begin{table*}[t]
\centering
\caption{\footnotesize Baseline demographic and clinical characteristics of patients across the internal and external cohorts. Continuous variables are presented as mean $\pm$ standard deviation (or median [IQR], as appropriate), and categorical variables are presented as number (\%). P-values compare cohorts using ANOVA or Kruskal--Wallis tests for continuous variables and $\chi^2$ test or Fisher's exact test for categorical variables, as appropriate.}
\label{tab:demographics}
\resizebox{0.8\textwidth}{!}{
\begin{tabular}{lccccc}
\toprule
\textbf{Characteristic} &
\textbf{Total} &
\textbf{Mayo Internal} &
\textbf{MIMIC External} &
\textbf{Mayo ED External} &
\textbf{P-value} \\
\midrule

Number of patients & 22,573 & 15,802 & 4,616 & 2,155 & $p < 0.001$ \\
\midrule
\rowcolor{gray!15}
\textbf{Demographics} &&&&&\\

Age (years) & 63.14 ± 17.34 & 64.16 ± 18.01 & 61.84 ± 17.33 & 58.38 ± 9.65 & -- \\

Female sex, n (\%) & 9,611 (42.6\%) & 6,454 (40.6\%) & 2,343 (50.8\%) & 832 (38.6\%) & $p < 0.001$ \\

Male sex, n (\%) & 13,001 (57.6\%)
  & 9,431 (59.3\%) & 2,272 (49.2\%) & 1,323 (61.4\%) & $p < 0.001$ \\

\rowcolor{gray!08}
\textbf{Race, n (\%)} &&&&&\\
\hspace{5mm} White & 14,564 (64.4\%, n=19,011) & 10,265 (83.9\%, n=12,240) & 3,043 (65.9\%) & 1,601 (74.3\%) & $p < 0.001$ \\

\hspace{5mm} Black or African American & 1,370 (6.1\%, n=19,011) & 474 (3.9\%, n=12,240) & 849 (18.4\%) & 52 (2.4\%) & $p < 0.001$ \\

\hspace{5mm} Asian & 460 (2.0\%, n=19,011) & 236 (1.9\%, n=12,240) & 168 (3.6\%) & 56 (2.6\%) & $p < 0.001$  \\

\hspace{5mm} American Indian or Alaska Native & 120 (0.5\%, n=19,011) & 111 (0.9\%, n=12,240) & 3 (0.1\%) & 6 (0.3\%) & $p < 0.001$ \\


\hspace{5mm} Other/Multiple races or Unknown/Not reported
& 6,099 (27.0\%, n=19,011) & 1,197 (9.8\%, n=12,240) & 553 (12.0\%) & 440 (20.4\%) & $p < 0.001$ \\

\rowcolor{gray!08}
\textbf{Ethnicity, n (\%)} &&&&&\\
\hspace{5mm} Hispanic or Latino & 859 (3.8\%, n=19,011) & 265 (2.2\%, n=12,240) & 253 (5.5\%) & 86 (4.0\%) & $p < 0.001$ \\

\hspace{5mm} Non-Hispanic or Latino & 16,063 (71.0\%, n=19,011) & 10276 (84\%, n=12,240) & 4,242 (91.9\%) & 1,596 (73.6\%) & $p < 0.001$  \\

\hspace{5mm} Unknown/Not reported & 5,709 (25.2\%, n=19,011) & 1482 (12.1\%, n=12,240) & 121 (2.6\%) & 483 (22.4\%) & $p < 0.001$  \\

\midrule
\rowcolor{gray!15}
\textbf{Comorbidities} &&&&&\\

Hypertension, n (\%) & 15,040 (66.5\%) & 11,664 (73.8\%) & 3,344 (72.4\%) & 32 (1.5\%) & $p=0.0060$ \\

Diabetes mellitus, n (\%) & 6,539 (28.9\%) & 4897 (31\%) & 1,623 (35.2\%) & 19 (0.9\%) & $p < 0.001$ \\

Hyperlipidemia, n (\%) & 12,651 (61.8\%, n=20,418) & 10,070 (63.7\%) & 2,581 (55.9\%) &-- & $p < 0.001$ \\

Coronary artery disease, n (\%) & 7,076 (43.3\%, n=20,418) & 9,707 (61.4\%) & 1,516 (32.8\%) & -- & $p < 0.001$ \\


Atrial fibrillation, n (\%) & 6,021 (36.9\%, n=16,329) & 4,775 (40.8\%, n=11,713) & 1,246 (27.0\%) & -- & $p < 0.001$ \\

Chronic kidney disease, n (\%) & 7,295 (32.2\%, n=20,418) & 5,911 (37.4\%) & 1,378 (29.9\%) & 6 (0.3\%) & $p < 0.001$ \\

Chronic obstructive pulmonary disease, n (\%) & 3,895 (19.0\%, n=20,418) & 3,060 (19.3\%) & 835 (18.1\%) & -- & $p=0.0697$ \\


\midrule
\rowcolor{gray!15}
\textbf{Clinical Outcomes Within 1 Year} &&&&&\\

Myocardial infarction, n (\%) & 655 (3.6\%) & 496 (3.1\%) & -- & 159 (7.4\%) & $p < 0.001$ \\

Heart failure event, n (\%) & 1,182 (6.6\%) & 1,010 (6.4\%) & -- & 172 (8\%) & $p < 0.001$ \\

Stroke, n (\%) & 779 (4.3\%) & 607 (3.8\%) & -- & 172 (8\%) & $p < 0.001$ \\

Death, n (\%) & 2,383 (13.3\%) & 2,338 (14.8\%) & 455 (9.9\%) & 45 (2.1\%) & $p < 0.001$ \\

\bottomrule
\end{tabular}
}
\end{table*}

\subsubsection{Data processing and Implementation}
Clinical notes were matched to chest X-ray studies using the closest document creation time from the ordering physician, retaining only notes created on or before the chest X-ray date. Clinical note text was summarized using a open source finetuned Qwen based LLM medical summarization model\footnote{Qwen-based LLM medical summarization model: \url{https://huggingface.co/mradermacher/Qwen2.5-7B-Instruct-medical_summary_latest-GGUF}.} implemented with llama-cpp. The model generated 3–4 English sentences covering available demographics, social and clinical history, medications, reason for visit, and symptoms. Missing or empty notes were excluded.

All models were implemented in PyTorch using the Hugging Face Transformers, PEFT, and Accelerate libraries. The proposed framework was developed with Python 3.11.15 and PyTorch 2.6.0+cu118, and trained on four NVIDIA RTX A6000 GPUs using bfloat16 precision. Images were processed using each VLM's native dynamic-resolution processor, and clinical histories were incorporated through multimodal instruction prompts. Additional materials are provided in the Supplemental. Source code and pretrained weights are publicly available\footnote{Source code: \url{https://github.com/OrchidPi/cxr-causal-reinforcement-reasoning}; Model weights: \url{https://huggingface.co/OrchidPi/Lingshu-7B-causal-reinforcement-reasoning}.}

\subsection{Evaluation Protocol} We compared our framework with both unimodal and multimodal baselines. Unimodal models included DenseNet~\cite{ye2020weaklysupervisedlesionlocalization} based pretrained on CheXpert \cite{chambon2024chexpertplusaugmentinglarge} and MedCLIP-ViT \cite{wang2022medclip} for CXR-only prediction, and BioGPT \cite{luo2022biogpt}, MedCLIP-ClinicalBERT\cite{wang2022medclip} for clinical-text-only prediction. For multimodal learning, we evaluated MedGemma-4B-IT \cite{sellergren2025medgemma}, Lingshu-7B \cite{xu2025lingshu}, Qwen3-VL-8B-Instruct \cite{qwen3technicalreport}, and Llama-3.2-11B-Vision-Instruct under three training paradigms: (1) \textbf{VLM Classification}, which directly predicts MACE from multimodal features; (2) \textbf{Single-VLM Multi-task Learning}, which jointly performs risk classification, event prediction, and reasoning generation using a shared decoder; and (3) \textbf{Proposed Role-Decoupled Multi-LLM framework}, which separates multimodal reasoning from downstream risk prediction and outcome generation. All VLMs were fine-tuned using the same parameter-efficient strategy, updating only the multimodal projection module and LoRA adapters while freezing the remaining pretrained parameters.

Separate token masks are constructed for \texttt{selected choices} and \texttt{selected events}, producing the objective: \(
    \mathcal{L}_{\mathrm{single}}
    =
    \lambda_{\mathrm{cls}}\mathcal{L}_{\mathrm{choice}}
    +
    \lambda_{\mathrm{gen}}\mathcal{L}_{\mathrm{event}},
\)
where both terms are autoregressive token-level cross-entropy losses. Only the risk decision and event prediction tokens were supervised, whereas the subsequent free-text reasoning was generated without direct supervision. This baseline evaluates whether a single VLM decoder can jointly learn risk prediction, structured event generation, and clinical reasoning. We further compared two multimodal settings: Direct VLM, in which MACE prediction and event generation were formulated as text-generation tasks, and VLM--BioGPT, in which the VLM-generated reasoning and projected multimodal features were provided to BioGPT for downstream risk classification and outcome-phrase generation. All VLMs were fine-tuned using the same parameter-efficient strategy, updating only the multimodal projection module and LoRA adapters while freezing the remaining pretrained parameters.

The Causal-RL model was initialized from the supervised VLM–BioGPT model. We split the internal training set into 80\% for Stage-I training of the VLM–BioGPT baseline and 20\% for Stage-II causal RL optimization. Our hypothesis was that causal RL could further improve reasoning quality by leveraging a smaller subset of unseen training data to refine the model’s decision policy. Its context-aware policy used a 16-entry EMA codebook with 32-dimensional codes and selected both a multimodal token-retention mask and a reasoning-length budget. The available reasoning budgets were 96 and 128 tokens. The policy was optimized using GRPO with 8 rollouts per study, a learning rate of $3\times10^{-4}$, and a clipping coefficient of 0.2. The rollout reward combined MACE prediction error, retained token proportion, and outcome-generation loss with weights of 1.0, 0.1, and 0.5, respectively. During inference, token-retention and reasoning-length actions were selected deterministically before final VLM reasoning and BioGPT classification.

\subsection{Evaluation Metrics}
Prediction performance was evaluated using AUROC, positive predictive value (PPV), and negative predictive value (NPV) with 95\% confidence intervals (Table~\ref{tab:model_performance_ci}). Reasoning quality was evaluated using Qwen2.5-32B-Instruct~\cite{qwen2.5} as an LLM-based judge by comparing generated reasoning with clinical histories (prompt provided in Supplementary Materials). Without ground-truth reasoning annotations, clinically relevant facts were categorized as matched ($M$), missing ($S$), mischaracterized ($C$), or hallucinated ($H$). We reported factual correctness (GREEN score: $M/(M+C+H)$), reasoning completeness ($M/(M+S+C+H)$), and hallucination rate ($H/(M+C+H)$) (Table~\ref{tab:green}). CXR-derived findings were excluded from these metrics. Clinical utility was assessed using prediction justification, clinical actionability, and expected calibration error (ECE)~\cite{murphy1967verification}, evaluating consistency between generated reasoning, predicted risk, clinical outcomes, and probability calibration (Table~\ref{tab:green}).

\begin{figure}[ht!]
\centering
\begin{subfigure}[b]{0.48\textwidth}
        \centering
        \includegraphics[width=\textwidth]{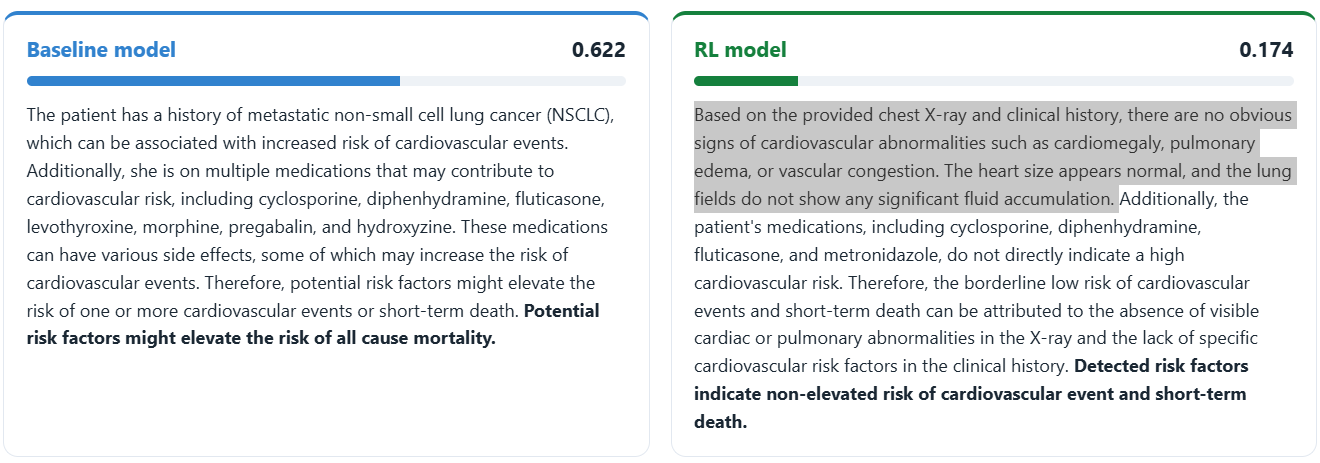}
        \caption{}
        \label{fig:sub2}
    \end{subfigure}
\begin{subfigure}[b]{0.48\textwidth}
        \centering
        \includegraphics[width=\linewidth]{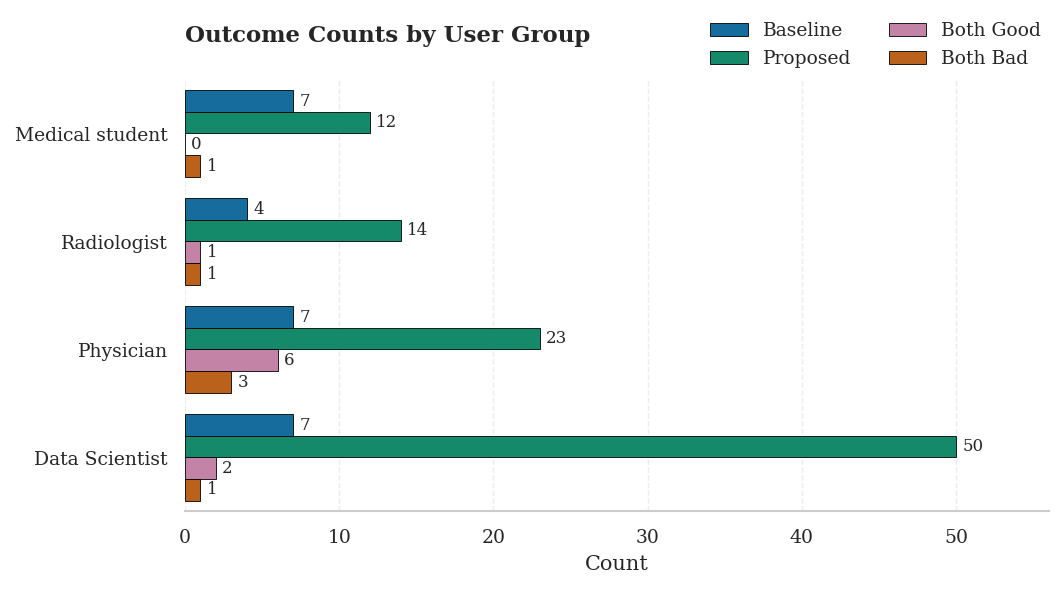}
        \caption{}
        \label{fig:sub2}
    \end{subfigure}

\caption{\footnotesize Reasoning quality evaluation. (a) Example true-negative case comparing baseline and proposed model reasoning. The baseline model incorrectly predicted high risk without considering relevant CXR findings, whereas the proposed RL-based model correctly predicted low risk with clinically grounded reasoning. (b) Expert preference comparison between baseline and proposed model reasoning across user groups.}
\label{fig:user_model}
\end{figure}

An expert preference study was performed on 20 randomly selected cases with blinded randomized comparison between the proposed and baseline reasoning. Experts from radiology, medicine, medical training, and AI/data science assessed reasoning preference and evidence support from CXR and clinical history (Table~\ref{tab:green}, Fig.~\ref{fig:user_model}). Statistical significance was evaluated using paired bootstrap resampling with 2,000 matched-case resamples.
Statistical significance was assessed using paired bootstrap resampling with 2,000 resamples of matched cases. For each resample, the difference between the Baseline and Proposed methods was recomputed. The two-sided $p$-value was calculated as $\min\!\left[1,\,2\min\!\left(P(\Delta \geq 0),P(\Delta \leq 0)\right)\right]$, where $\Delta$ denotes the Baseline--Proposed difference.

\section{Results}
\subsection{Quantitative Performance}
We compared single-modality models (image only and text only), multimodal baselines using one LLM for joint classification and reasoning generation, and two-LLM with and without causal RL. As shown in Table~\ref{tab:model_performance_ci}, image-only models performed near chance (AUC $\approx$ 0.5) across the three datasets, whereas text-only models achieved substantially better discrimination, with the MedCLIP text encoder reaching AUCs of 0.743, 0.798, and 0.804 on the internal, MIMIC-external, and ED-external datasets, respectively. After incorporating both image and clinical history, multimodal models improved further, but the one-LLM baseline remained limited, especially on the internal and MIMIC-external datasets. In contrast, while using the same pre-trained LLM backbone, our proposed two-LLM model with causal RL consistently maintained competitive performance and yielded modest gains over the one-LLM baseline, while also improving or matching the two-LLM baseline without causal RL on most settings. Compared with the model without causal RL, the proposed approach showed modest improvement on the internal dataset and a larger gain on the ED external cohort, where Lingshu-based AUC increased from 0.804 to 0.843. Performance on the MIMIC external cohort was largely unchanged in terms of AUC. For PPV, the causal RL model generally outperformed the other two models, increasing from 0.241 and 0.247 to 0.279. 

For clinical reasoning evaluation, we randomly sampled 1,000 cases from each dataset, stratified across low-, borderline low-, borderline high-, and high-risk prediction prompts. Model-generated reasoning for these cases was evaluated using an LLM-based judge to minimize bias arising from differences in prompt wording across risk categories. To verify the reliability of the automated evaluation, we manually reviewed cases in which the LLM judge appeared to incorrectly identify hallucinations. The baseline model employed a single LLM for joint risk classification and reasoning generation, whereas the proposed framework used a role-decoupled dual-LLM architecture with causal reinforcement learning, separating reasoning generation from downstream classification. As shown in Table~\ref{tab:green}, the proposed model improved overall reasoning quality on the Mayo internal cohort, with the GREEN score increasing from $0.405 \pm 0.453$ to $0.547 \pm 0.425$ ($p < 0.001$), and showed a larger gain on the Mayo ED external cohort, from $0.269 \pm 0.406$ to $0.660 \pm 0.392$ ($p < 0.001$). Performance on the MIMIC external cohort was similar on overall GREEN score. Beyond the overall score, the proposed model achieved significant improvements in several clinical utility measures, including prediction justification, calibration error, and expert preference, with many comparisons reaching $p < 0.001$.

\begin{figure}[t]
\centering
\includegraphics[width=0.80\linewidth]{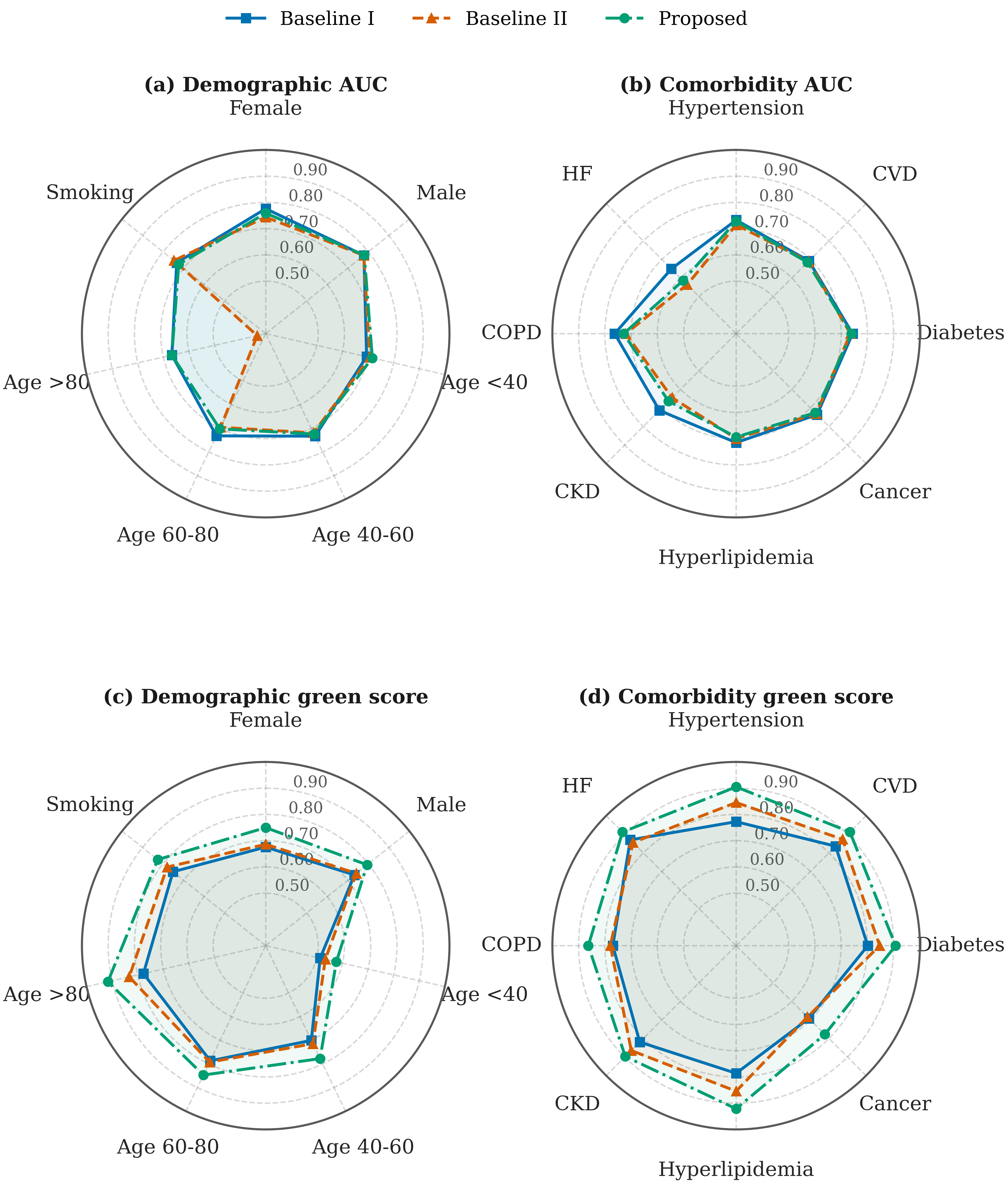}
\caption{\footnotesize Subgroup analysis on External MIMIC datasets comparing Three models across AUC (above) and average green score (below).}
\label{fig:mimic_subgroup}
\end{figure}

Figure~\ref{fig:mimic_subgroup} summarizes subgroup performance on the external MIMIC cohort across demographic and comorbidity strata. Baseline I employs a single LLM for joint classification and reasoning generation, Baseline II uses a role-decoupled dual-LLM architecture without causal RL, and the proposed model incorporates causal RL. Although all three models achieve comparable AUC across subgroups, the proposed framework consistently attains higher and more stable GREEN scores. These results suggest that causal RL improves the faithfulness and consistency of clinical reasoning across diverse patient populations, reducing subgroup-dependent variability in generated explanations while maintaining predictive performance. This more uniform reasoning behavior has the potential to mitigate representational bias in AI-generated clinical explanations, thereby improving the reliability and fairness of decision support across patient subgroups.

\begin{table*}[htb!]
\centering
\caption{Performance comparison of single modality and multimodal models across internal and external datasets. Values are reported as point estimate $\pm$ half-width of the 95\% confidence interval.}
\label{tab:model_performance_ci}
\tiny
\setlength{\tabcolsep}{2pt}
\renewcommand{\arraystretch}{1.08}
\resizebox{\textwidth}{!}{%
\begin{tabular}{llccccccccc}
\toprule
\textbf{} & \textbf{Models}
& \multicolumn{3}{c}{\textbf{AUC}}
& \multicolumn{3}{c}{\textbf{NPV}}
& \multicolumn{3}{c}{\textbf{PPV}} \\
\cmidrule(lr){3-5} \cmidrule(lr){6-8} \cmidrule(lr){9-11}
& 
& \textbf{Mayo-Internal}
& \textbf{MIMIC-External}
& \textbf{Mayo ED-External}
& \textbf{Mayo-Internal}
& \textbf{MIMIC-External}
& \textbf{Mayo ED-External}
& \textbf{Mayo-Internal}
& \textbf{MIMIC-External}
& \textbf{Mayo ED-External} \\
\midrule

\rowcolor{gray!15}
\multicolumn{11}{l}{\textbf{Single modality without reasoning}} \\
& Image only (CheXpert-DenseNet)
& 0.500 $\pm$ 0.004 & 0.526 $\pm$ 0.003 & 0.499 $\pm$ 0.007
& 0.684 $\pm$ 0.005 & 0.798 $\pm$ 0.003 & 0.883 $\pm$ 0.003
& 0.322 $\pm$ 0.005 & 0.234 $\pm$ 0.003 & 0.100 $\pm$ 0.002 \\
& Image only (MedCLIP-ViT)
& 0.539 $\pm$ 0.003 & 0.520 $\pm$ 0.003 & 0.489 $\pm$ 0.013
& 0.712 $\pm$ 0.004 & 0.789 $\pm$ 0.004 & 0.932 $\pm$ 0.003
& 0.347 $\pm$ 0.005 & 0.225 $\pm$ 0.003 & 0.051 $\pm$ 0.002 \\
& Text only (BioGPT)
& 0.740 $\pm$ 0.003 & 0.772 $\pm$ 0.004 & \textbf{0.808 $\pm$ 0.007}
& 0.797 $\pm$ 0.003 & 0.893 $\pm$ 0.003 & \textbf{0.961 $\pm$ 0.002}
& \textbf{0.518 $\pm$ 0.004} & 0.416 $\pm$ 0.005 & \textbf{0.315 $\pm$ 0.006} \\
& Text only (MedCLIP-ClinicalBERT)
& \textbf{0.743 $\pm$ 0.003} & \textbf{0.798 $\pm$ 0.004} & 0.804 $\pm$ 0.005
& \textbf{0.803 $\pm$ 0.004} & 0.903 $\pm$ 0.002 & 0.960 $\pm$ 0.002
& 0.515 $\pm$ 0.004 & \textbf{0.425 $\pm$ 0.006} & 0.294 $\pm$ 0.006 \\

\midrule
\rowcolor{gray!15}
\multicolumn{11}{l}{\textbf{Multimodal with reasoning}} \\
\rowcolor{gray!08}
\multicolumn{11}{l}{\textbf{Medical domain MLLM: MedGemma-4B-IT}} \\
& One LLM cls + reasoning
& 0.632 $\pm$ 0.003 & 0.705 $\pm$ 0.005 & 0.768 $\pm$ 0.013
& 0.749 $\pm$ 0.003 & 0.871 $\pm$ 0.003 & 0.961 $\pm$ 0.003
& 0.427 $\pm$ 0.003 & 0.324 $\pm$ 0.006 & 0.200 $\pm$ 0.006 \\
& Two LLM cls + reasoning (stepI)
& 0.730 $\pm$ 0.003 & \textbf{0.774 $\pm$ 0.003} & \textbf{0.842 $\pm$ 0.013}
& 0.796 $\pm$ 0.002 & 0.895 $\pm$ 0.004 & \textbf{0.964 $\pm$ 0.002}
& 0.501 $\pm$ 0.005 & \textbf{0.385 $\pm$ 0.005} & \textbf{0.325 $\pm$ 0.011} \\
& \textit{Proposed: Two LLM cls + reasoning w/ Causal RL (stepII)}
& \textbf{0.733 $\pm$ 0.004} & 0.752 $\pm$ 0.007 & 0.834 $\pm$ 0.006
& \textbf{0.806 $\pm$ 0.005} & \textbf{0.896 $\pm$ 0.003} & 0.963 $\pm$ 0.002
& \textbf{0.503 $\pm$ 0.004} & 0.364 $\pm$ 0.010 & 0.322 $\pm$ 0.007 \\

\rowcolor{gray!08}
\multicolumn{11}{l}{\textbf{Medical domain MLLM: Lingshu-7B}} \\
& One LLM cls + reasoning
& 0.652 $\pm$ 0.003 & 0.764 $\pm$ 0.004 & \textbf{0.855 $\pm$ 0.004}
& 0.764 $\pm$ 0.003 & \textbf{0.901 $\pm$ 0.003} & \textbf{0.974 $\pm$ 0.002}
& 0.427 $\pm$ 0.004 & 0.378 $\pm$ 0.006 & 0.263 $\pm$ 0.006 \\
& Two LLM cls + reasoning (stepI)
& 0.718 $\pm$ 0.003 & \textbf{0.762 $\pm$ 0.004} & 0.804 $\pm$ 0.007
& \textbf{0.801 $\pm$ 0.003} & 0.897 $\pm$ 0.003 & 0.956 $\pm$ 0.004
& 0.492 $\pm$ 0.003 & \textbf{0.387 $\pm$ 0.005} & 0.233 $\pm$ 0.007 \\
& \textit{Proposed: Two LLM cls + reasoning w/ Causal RL (stepII)}
& \textbf{0.720 $\pm$ 0.003} & 0.760 $\pm$ 0.005 & 0.843 $\pm$ 0.008
& 0.790 $\pm$ 0.004 & 0.897 $\pm$ 0.003 & 0.962 $\pm$ 0.002
& \textbf{0.507 $\pm$ 0.004} & 0.380 $\pm$ 0.006 & \textbf{0.293 $\pm$ 0.014} \\

\rowcolor{gray!08}
\multicolumn{11}{l}{\textbf{General domain MLLM: Qwen3-VL-8B-Instruct}} \\
& One LLM cls + reasoning
& 0.625 $\pm$ 0.003 & 0.700 $\pm$ 0.004 & 0.779 $\pm$ 0.007
& 0.744 $\pm$ 0.004 & 0.868 $\pm$ 0.003 & 0.949 $\pm$ 0.002
& 0.418 $\pm$ 0.004 & 0.336 $\pm$ 0.006 & 0.230 $\pm$ 0.006 \\
& Two LLM cls + reasoning (stepI)
& 0.711 $\pm$ 0.002 & \textbf{0.756 $\pm$ 0.003} & \textbf{0.849 $\pm$ 0.006}
& 0.795 $\pm$ 0.002 & \textbf{0.891 $\pm$ 0.002} & 0.962 $\pm$ 0.002
& \textbf{0.483 $\pm$ 0.003} & \textbf{0.385 $\pm$ 0.004} & 0.292 $\pm$ 0.008 \\
& \textit{Proposed: Two LLM cls + reasoning w/ Causal RL (stepII)}
& \textbf{0.711 $\pm$ 0.004} & 0.747 $\pm$ 0.004 & 0.845 $\pm$ 0.005
& \textbf{0.799 $\pm$ 0.003} & 0.887 $\pm$ 0.003 & \textbf{0.963 $\pm$ 0.002}
& 0.477 $\pm$ 0.004 & 0.382 $\pm$ 0.004 & \textbf{0.294 $\pm$ 0.006} \\

\rowcolor{gray!08}
\multicolumn{11}{l}{\textbf{General domain MLLM: Llama-3.2-11B-Vision-Instruct}} \\
& One LLM cls + reasoning
& 0.684 $\pm$ 0.004 & 0.691 $\pm$ 0.004 & \textbf{0.826 $\pm$ 0.009}
& 0.784 $\pm$ 0.004 & 0.872 $\pm$ 0.003 & 0.960 $\pm$ 0.003
& 0.450 $\pm$ 0.005 & 0.312 $\pm$ 0.005 & 0.241 $\pm$ 0.006 \\
& Two LLM cls + reasoning (stepI)
& 0.702 $\pm$ 0.003 & \textbf{0.759 $\pm$ 0.006} & 0.791 $\pm$ 0.008
& 0.780 $\pm$ 0.003 & 0.887 $\pm$ 0.004 & 0.957 $\pm$ 0.002
& \textbf{0.487 $\pm$ 0.004} & \textbf{0.388 $\pm$ 0.008} & 0.247 $\pm$ 0.007 \\
& \textit{Proposed: Two LLM cls + reasoning w/ Causal RL (stepII)}
& \textbf{0.710 $\pm$ 0.003} & 0.752 $\pm$ 0.003 & 0.809 $\pm$ 0.007
& \textbf{0.792 $\pm$ 0.002} & \textbf{0.888 $\pm$ 0.002} & \textbf{0.960 $\pm$ 0.002}
& 0.482 $\pm$ 0.004 & 0.377 $\pm$ 0.006 & \textbf{0.279 $\pm$ 0.006} \\

\bottomrule
\end{tabular}%
}
\end{table*}

\begin{table*}[htb!]
\centering
\caption{Evaluation of multimodal clinical reasoning quality and utility. Reasoning quality metrics are reported as mean $\pm$ standard deviation, while clinical utility metrics are reported as point estimate $\pm$ half-width of the 95\% confidence interval. `--' means not applicable.}
\label{tab:green}
\scriptsize
\setlength{\tabcolsep}{3pt}
\renewcommand{\arraystretch}{1.08}
\resizebox{\textwidth}{!}{%
\begin{tabular}{lccccccccc}
\toprule
\textbf{Evaluation Metric}
& \multicolumn{3}{c}{\textbf{Mayo Internal}}
& \multicolumn{3}{c}{\textbf{MIMIC External}}
& \multicolumn{3}{c}{\textbf{Mayo ED External}} \\
\cmidrule(lr){2-4} \cmidrule(lr){5-7} \cmidrule(lr){8-10}
& \textbf{Baseline}
& \textbf{Proposed}
& \textbf{P-value}
& \textbf{Baseline}
& \textbf{Proposed}
& \textbf{P-value}
& \textbf{Baseline}
& \textbf{Proposed}
& \textbf{P-value} \\
\midrule


\rowcolor{gray!15}
\multicolumn{10}{l}{\textbf{Reasoning Quality}} \\

Factual correctness
& 0.405 $\pm$ 0.453 & 0.547 $\pm$ 0.425 & $p < 0.001$
& 0.692 $\pm$ 0.341 & 0.670 $\pm$ 0.385 & $p = 0.404$
& 0.269 $\pm$ 0.406 & 0.660 $\pm$ 0.392 & $p < 0.001$ \\

Reasoning completeness
& 0.194 $\pm$ 0.297 & 0.321 $\pm$ 0.312 & $p < 0.001$
& 0.368 $\pm$ 0.260 & 0.285 $\pm$ 0.227 & $p < 0.001$
& 0.091 $\pm$ 0.202 & 0.364 $\pm$ 0.285 & $p < 0.001$ \\

Hallucination $\downarrow$
& 0.587 $\pm$ 0.456 & 0.447 $\pm$ 0.424 & $p < 0.001$
& 0.302 $\pm$ 0.341 & 0.326 $\pm$ 0.383 & $p = 0.329$
& 0.724 $\pm$ 0.411 & 0.337 $\pm$ 0.390 & $p < 0.001$ \\

\rowcolor{gray!15}
\multicolumn{10}{l}{\textbf{Clinical Utility}} \\

Prediction justification
& 0.249 ± 0.014 & 0.994 $\pm$ 0.003 & $p < 0.001$
& 0.266 ± 0.013 & 0.997 $\pm$ 0.002 & $p < 0.001$
& 0.346 ± 0.015 & 0.995 $\pm$ 0.002 & $p < 0.001$ \\

Clinical actionability
& 0.621 ± 0.032 & 0.638 $\pm$ 0.015 & $p = 0.168$
& 0.306 ± 0.028 & 0.700 $\pm$ 0.014 & $p < 0.001$
& 0.904 ± 0.015 & 0.781 $\pm$ 0.013 & $p < 0.001$ \\

Expected Calibration Error (ECE) $\downarrow$
& 0.107 ± 0.016 & 0.031 $\pm$ 0.010 & $p = 0.009$
& 0.436 ± 0.011 & 0.095 $\pm$ 0.010 & $p < 0.001$
& 0.163 ± 0.009 & 0.117 $\pm$ 0.008 & $p < 0.001$ \\


\rowcolor{gray!15}
\multicolumn{10}{l}{\textbf{Expert Preference}}\\
Percentage & -- & -- & --
& 32.76\% & 67.24\% & 0.012
& 9.09\% & 90.91\% & $ p < 0.001$ \\


\bottomrule
\end{tabular}%
}
\end{table*}

\subsection{Ablation Study}
We conducted an ablation study of the proposed role-decoupled dual-LLM architecture and causal RL using the best-performing Lingshu-7B model. Performance on both the internal and external cohorts is summarized in Table~\ref{tab:ablation_lingshu}.In addition to the baseline and proposed models, we evaluated seven ablation settings to quantify the contribution of individual modalities, architectural components, and training strategies: (1) \textbf{Modality contribution:} \emph{Image-only} removes the clinical context and uses only the CXR and prompt as input, while \emph{Text-only} removes the CXR and retains only the clinical context. (2) \textbf{Representation learning:} \emph{No Text Adapter} removes the text adapter used for multimodal fusion, and \emph{Frozen Vision Merger} freezes the vision merger during fine-tuning - restricts relevant learning from CXR. (3) \textbf{Reasoning supervision:}  \emph{No Risk Phrase Generation} removes the BioGPT-based risk phrase generation module, while \emph{Classification-only} performs MACE classification without generating reasoning. (4) \textbf{Model architecture:} \emph{Frozen Reasoning LLM} freezes the first-stage LLM decoder during training, evaluating the contribution of adapting the reasoning model.

As shown in Table~\ref{tab:ablation_lingshu}, for Mayo-Internal, removing the text adapter causes the largest degradation relative to the two-LLM baseline, reducing AUC, NPV, and PPV from $0.718$, $0.801$, and $0.492$ to $0.639$, $0.746$, and $0.437$, respectively. Removing either input modality also degrades performance: excluding text lowers AUC by $0.036$, whereas excluding images lowers it by $0.025$. Freezing the projector similarly reduces AUC to $0.707$, while freezing the decoder has a smaller effect ($0.715$), suggesting that projector adaptation is particularly important for multimodal alignment.

The external evaluations further demonstrate the importance of textual information. On MIMIC-External, removing text or the text adapter reduces AUC from the Baseline value of $0.762$ to $0.638$ and $0.719$, respectively. Freezing the decoder also decreases AUC to $0.744$. Although removing the image modality achieves a higher MIMIC Green Score than the Proposed model ($0.832$ vs.\ $0.776$), its Mayo ED AUC decreases substantially from $0.843$ to $0.752$, indicating weaker cross-domain robustness.

Finally, causal RL improves the model most clearly on Mayo ED, increasing AUC from $0.804$ to $0.843$ and PPV from $0.233$ to $0.293$, while also improving the MIMIC Green Score from $0.715$ to $0.776$. Using only one LLM generally weakens PPV and prevents generation-quality evaluation, supporting the use of separate classification and reasoning stages. The results indicate that text-adapter learning, multimodal inputs, and causal RL play complementary roles: the text pathway drives discrimination, trainable cross-modal components support adaptation, and causal RL improves the balance between predictive accuracy and clinically coherent reasoning.


\begin{table*}[htb!]
\centering
\caption{Ablation study of Lingshu-7B across internal and external datasets. `--' means not applicable}
\label{tab:ablation_lingshu}
\scriptsize
\setlength{\tabcolsep}{3pt}
\renewcommand{\arraystretch}{1.05}
\resizebox{\textwidth}{!}{%
\begin{tabular}{lcccccccccccc}
\toprule
\textbf{Ablation}
& \multicolumn{4}{c}{\textbf{Mayo-Internal}}
& \multicolumn{4}{c}{\textbf{MIMIC-External}}
& \multicolumn{4}{c}{\textbf{Mayo ED-External}} \\
\cmidrule(lr){2-5} \cmidrule(lr){6-9} \cmidrule(lr){10-13}
\textbf{Lingshu-7B}
& \textbf{AUC} & \textbf{NPV} & \textbf{PPV} & \textbf{Green Score}
& \textbf{AUC} & \textbf{NPV} & \textbf{PPV} & \textbf{Green Score}
& \textbf{AUC} & \textbf{NPV} & \textbf{PPV} & \textbf{Green Score} \\
\midrule

\rowcolor{gray!15}
\multicolumn{13}{l}{\textbf{Modality contribution}} \\
\emph{Image-only}
& 0.682 $\pm$ 0.006 & 0.784 $\pm$ 0.004 & 0.471 $\pm$ 0.006 & --
& 0.638 $\pm$ 0.004 & 0.852 $\pm$ 0.003 & 0.305 $\pm$ 0.004 & --
& 0.750 $\pm$ 0.007 & 0.949 $\pm$ 0.003 & 0.206 $\pm$ 0.006 & -- \\

\emph{Text-only}
& 0.693 $\pm$ 0.003 & 0.788 $\pm$ 0.004 & 0.466 $\pm$ 0.003 & \textbf{0.577 $\pm$ 0.416}
& 0.762 $\pm$ 0.003 & 0.897 $\pm$ 0.001 & 0.387 $\pm$ 0.004 & \textbf{0.832 $\pm$ 0.282}
& 0.752 $\pm$ 0.009 & 0.946 $\pm$ 0.002 & 0.238 $\pm$ 0.009 & \textbf{0.759 $\pm$ 0.310} \\
\rowcolor{gray!15}
\multicolumn{13}{l}{\textbf{Representation learning}} \\
\emph{No Text Adapter}
& 0.639 $\pm$ 0.002 & 0.746 $\pm$ 0.003 & 0.437 $\pm$ 0.003 & 0.429 $\pm$ 0.364
& 0.719 $\pm$ 0.007 & 0.874 $\pm$ 0.004 & 0.367 $\pm$ 0.007 & 0.728 $\pm$ 0.296
& 0.733 $\pm$ 0.006 & 0.944 $\pm$ 0.002 & 0.208 $\pm$ 0.007 & 0.626 $\pm$ 0.309 \\

\emph{Frozen Vision Merger}
& 0.707 $\pm$ 0.005 & 0.784 $\pm$ 0.003 & 0.489 $\pm$ 0.007 & 0.429 $\pm$ 0.364
& 0.760 $\pm$ 0.003 & 0.894 $\pm$ 0.002 & 0.389 $\pm$ 0.004 & 0.729 $\pm$ 0.296
& 0.811 $\pm$ 0.006 & 0.960 $\pm$ 0.003 & 0.225 $\pm$ 0.006 & 0.627 $\pm$ 0.308 \\

\rowcolor{gray!15}
\multicolumn{13}{l}{\textbf{Reasoning supervision}} \\
\emph{No Risk Phrase Generation}
& 0.709 $\pm$ 0.003 & 0.790 $\pm$ 0.004 & 0.485 $\pm$ 0.003 & 0.440 $\pm$ 0.364
& \textbf{0.765 $\pm$ 0.003} & 0.898 $\pm$ 0.002 & \textbf{0.394 $\pm$ 0.003} & 0.727 $\pm$ 0.288
& 0.789 $\pm$ 0.007 & 0.951 $\pm$ 0.003 & 0.261 $\pm$ 0.007 & 0.634 $\pm$ 0.308 \\

\emph{Classification Only}
& 0.715 $\pm$ 0.003 & 0.788 $\pm$ 0.003 & 0.494 $\pm$ 0.005 & --
& 0.753 $\pm$ 0.004 & 0.898 $\pm$ 0.003 & 0.369 $\pm$ 0.006 & --
& 0.770 $\pm$ 0.007 & 0.949 $\pm$ 0.003 & 0.237 $\pm$ 0.006 & -- \\

\rowcolor{gray!15}
\multicolumn{13}{l}{\textbf{Model architecture}} \\
\emph{Frozen Reasoning LLM}
& 0.715 $\pm$ 0.004 & \textbf{0.795 $\pm$ 0.004} & 0.500 $\pm$ 0.003 & 0.442 $\pm$ 0.364
& 0.744 $\pm$ 0.003 & 0.891 $\pm$ 0.002 & 0.364 $\pm$ 0.003 & 0.726 $\pm$ 0.294
& 0.759 $\pm$ 0.006 & 0.950 $\pm$ 0.002 & 0.226 $\pm$ 0.005 & 0.625 $\pm$ 0.309 \\
\midrule
\emph{Baseline: One LLM cls + reasoning }
& 0.652 $\pm$ 0.003 & 0.764 $\pm$ 0.003 & 0.427 $\pm$ 0.004 & 0.395 $\pm$ 0.449 
& 0.764 $\pm$ 0.004 & \textbf{0.901 $\pm$ 0.003} & 0.378 $\pm$ 0.006 & 0.701 $\pm$ 0.338
& \textbf{0.855 $\pm$ 0.004} & \textbf{0.974 $\pm$ 0.002} & 0.263 $\pm$ 0.006 & 0.289 $\pm$ 0.422 \\


\emph{Proposed: Two LLM cls + reasoning w/ Causal RL}
& \textbf{0.720 $\pm$ 0.003} & 0.790 $\pm$ 0.004 & \textbf{0.507 $\pm$ 0.004} & \textbf{0.498 $\pm$ 0.402}
& 0.760 $\pm$ 0.005 & 0.897 $\pm$ 0.003 & 0.380 $\pm$ 0.006 & \textbf{0.776 $\pm$ 0.297}
& 0.843 $\pm$ 0.008 & 0.962 $\pm$ 0.002 & \textbf{0.293 $\pm$ 0.014} & 0.616 $\pm$ 0.345 \\

\bottomrule
\end{tabular}%
}
\end{table*}

\section{Discussion}
Our work advances the emerging field of multimodal clinical foundation models by addressing several limitations of current medical vision-language models. While recent LLMs and VLMs have demonstrated impressive performance on image interpretation, report generation, and visual question answering, they are primarily optimized for image-text alignment and language generation rather than prognostic prediction. Moreover, most existing approaches employ a single decoder to jointly perform reasoning and prediction, which can lead to competing optimization objectives and explanations that are weakly aligned with the final clinical decision. In contrast, our framework explicitly separates multimodal reasoning from downstream risk prediction through a role-decoupled dual-LLM architecture and optimizes reasoning using intervention-guided causal reinforcement learning. This design enables reasoning to be directly optimized for prediction while maintaining a differentiable connection between multimodal evidence and downstream classification.

Our findings further support the value of opportunistic cardiovascular screening from routinely acquired clinical data. Previous cardiovascular risk prediction models have largely relied on structured risk factors or dedicated cardiac imaging, whereas our framework demonstrates that chest radiographs and physician-authored clinical histories together provide complementary prognostic information for MACE prediction. The superior performance of multimodal models over image-only and text-only baselines indicates that integrating upstream clinical context with imaging more closely reflects physician decision-making and yields more informative patient representations than either modality alone.

Beyond predictive performance, our study demonstrates that explicitly optimizing reasoning improves explanation quality. Compared with conventional single-LLM architectures, the proposed framework consistently achieved higher GREEN scores and stronger expert preference across internal and external cohorts while maintaining competitive discrimination. These findings suggest that intervention-guided causal reinforcement learning improves the faithfulness of generated reasoning by encouraging the model to select clinically relevant evidence and suppress spurious multimodal associations. Notably, the improvement in GREEN scores was also consistent across demographic and comorbidity subgroups, suggesting reduced subgroup-dependent variability in generated explanations and highlighting the potential of reasoning optimization to improve the robustness and fairness of multimodal clinical decision support.

Several limitations should be considered. First, this was a retrospective study, and prospective evaluation is required to determine whether the generated reasoning improves clinician trust, workflow efficiency, or patient outcomes. Second, although external validation was performed on two independent cohorts, additional studies across diverse healthcare systems and imaging protocols are needed to establish generalizability. Third, the proposed causal reinforcement learning framework models interventions on the model's information pathway rather than causal relationships between clinical variables and outcomes; therefore, the generated explanations should be interpreted as evidence-supported reasoning rather than causal clinical inference. Finally, our framework uses physician-authored clinical histories available at the time of imaging, and performance may be affected by variations in documentation quality and completeness. Future work will extend the framework to incorporate additional longitudinal modalities, including laboratory results, medications, electrocardiograms, genomics, and wearable sensor data, and evaluate its clinical utility in prospective deployment studies.

\section*{REFERENCES}
\bibliographystyle{IEEEtran}
\bibliography{ref}





\end{document}